\documentclass[letterpaper, 10 pt, conference]{IEEEtran}  

\IEEEoverridecommandlockouts                              
\pdfoutput=1                                                    

\usepackage{epsfig} 
\usepackage{mathptmx} 
\usepackage{times} 
\usepackage{amsmath} 
\usepackage{amssymb}  
\usepackage{graphicx} 
\usepackage{graphics} 
\usepackage{grffile} 
\usepackage[T1]{fontenc}
\usepackage{amsfonts}
\usepackage{booktabs}
\usepackage{siunitx}
\usepackage{caption} 
\usepackage{booktabs}
\usepackage{graphicx}
\usepackage{algorithmic}
\usepackage[ruled]{algorithm2e}
\usepackage{float}

\usepackage{mwe,tikz}\usepackage[percent]{overpic}

\usepackage{subcaption}
\usepackage{array, tabularx, makecell, booktabs}%
\usepackage{geometry}
\usepackage{comment}

\title{\LARGE \bf
Connecting Deep-Reinforcement-Learning-based Obstacle Avoidance with Conventional Global Planners using Waypoint Generators
}

\author{Linh K{\"a}stner$^{1}$\thanks{$^{1}$Linh K{\"a}stner, Teham Buiyan, Xinlin Zhao, Zhengcheng Shen, Cornelius Marx and Jens Lambrecht are with the Chair Industry Grade Networks and Clouds, Faculty of Electrical Engineering, and Computer Science,				
		Berlin Institute of Technology, Berlin, Germany
		{\tt\small linhdoan@tu-berlin.de}}, Teham Buiyan$^{1}$, Xinlin Zhao$^{1}$, Zhengcheng Shen$^{1}$ \\Jens Lambrecht$^{1}$ and Cornelius Marx$^{1}$
}

\begin{document}

\maketitle
\thispagestyle{empty}
\pagestyle{empty}


\begin{abstract}

Deep Reinforcement Learning has emerged as an efficient dynamic obstacle avoidance method in highly dynamic environments. It has the potential to replace overly conservative or inefficient navigation approaches. However, the integration of Deep Reinforcement Learning into existing navigation systems is still an open frontier due to the myopic nature of Deep-Reinforcement-Learning-based navigation, which hinders its widespread integration into current navigation systems. In this paper, we propose the concept of an intermediate planner to interconnect novel Deep-Reinforcement-Learning-based obstacle avoidance with conventional global planning methods using waypoint generation. Therefore, we integrate different waypoint generators into existing navigation systems and compare the joint system against traditional ones. We found an increased performance in terms of safety, efficiency and path smoothness especially in highly dynamic environments.

\end{abstract}


\section{Introduction}
Recently, mobile robots have become important tools in the industry, especially in logistics \cite{fragapane2020increasing}, \cite{alatise2020review}. Typically, robotic navigation systems employ hierarchical planners consisting of a global planner, which calculates the optimal path using search-based approaches like A-star or Random Rapid Tree (RRT) search, and a local planner, which executes it considering local observations and unknown obstacles. While current systems can cope well with static environments, dynamic ones employing various dynamic obstacles still pose a great challenge. Deep Reinforcement Learning (DRL) has emerged as an end-to-end planning method to replace overly conservative planners in highly dynamic environments due to its ability to achieve efficient navigation and obstacle avoidance from raw sensor input. 
However, the integration of DRL into conventional planners and usage in industries is limited due to its myopic nature and tedious training. Furthermore, due to the lack of a long term memory, DRL-based methods are very sensitive to local minima and thus can easily be trapped in complex situations like long corridors, corners, or dead ends.
Efforts to increase the capabilities of DRL-based systems to achieve long-range-navigation are complex and intensify the already tedious training procedures \cite{faust2018prm}, \cite{chiang2019learning}. 

\begin{figure}[]
	\centering
	\includegraphics[width=2.65in, height=2.3in]{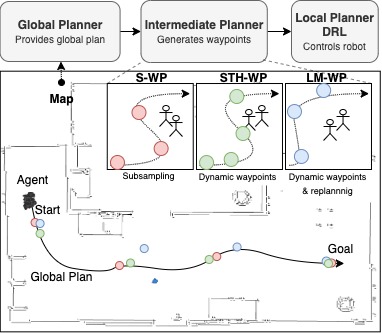}
	\caption{Our proposed hierarchical navigation system includes an intermediate planner which interconnects a traditional global planner and a DRL-based local planner. We propose three different waypoint generators based on: subsampling (SUB-WP), a spatial-time-horizon (STH-WP), and on landmarks (LM-WP).}
	\label{intro}
\end{figure}

In this work, we propose an interconnection entity called the intermediate planner to combine traditional global planning methods with a DRL-based local planner by using waypoints. Contrarily to existing works, the intermediate planner should generate waypoints more dynamically and flexibly along a given path while considering global as well as local information. This not only provides a shorter goal horizon but also adds an additional security layer in preparation for the local planner, which only has access to local sensor data. In particular, we propose two different waypoint generators that we compare against each other and against conventional navigation systems in terms of safety, robustness and efficiency.
The main contributions of this work are the following:
\begin{itemize}
    \item Proposal of two different waypoint generators to interconnect DRL-based local navigation with global methods and make DRL feasible for integration long-range-navigation.
    \item Integration of our approaches into the robot operating system (ROS) for a combination with conventional planners and to make deployment towards a real robot feasible.
    \item Extensive evaluation of the waypoint generators against one another as well as against conventional planning systems.

\end{itemize}
The paper is structured as follows. Sec. II begins with related works. Subsequently, the methodology is presented in Sec. III. Sec. IV presents the results and discussion. Finally, Sec. V will give a conclusion and outlook. We made the code publicly available under https://github.com/ignc-research/arena-rosnav.


\section{Related Works}
DRL has been extensively studied in various publications ranging from simulation, gaming, to navigation and achieved remarkable results in these areas.
Mnih et al. \cite{mnih2015human} first introduced the combination of Reinforcement Learning with deep neural networks and trained an agent to navigate in unknown environments based solely on sensor observations.
In recent years, the usage of DRL for the path planning and obstacle avoidance problem has intensified \cite{faust2018prm} - \cite{dugas2020navrep}. 
Chen et al. \cite{chen2017decentralized}, \cite{chen2017socially} proposed a DRL-based navigation for crowded environments and successfully transferred the approach to a real robot. The agent could contemplate to social behaviors and navigate robustly in cluttered environments. Everett et al. \cite{everett2018motion} extended the approach with a Long-Short-Term-Memory (LSTM) architecture \cite{hochreiter1997long} and made it feasible for a dynamic number of obstacles.
In \cite{dugas2020navrep}, Dugas et al. propose a platform consisting of a 2D simulation environment to train and test different DRL algorithms. The researchers integrate a variety of different DRL-based navigation approaches into their platform and compare them against one another. 
Faust et al. \cite{faust2018prm} combines a DRL-based local planner with sampling-based traditional planners. Global maps are generated by a DRL agent and, subsequently, the same DRL agent is utilized to execute the navigation. The researchers achieved long-range navigation in large scale office maps. However, the training is tedious, requires a high amount of parameters, and a not intuitive pre-setup.
Chen et al. \cite{chen2020learning}, \cite{chen2020soundspaces} utilizes DRL to generate waypoints by considering audio visual clues and an acoustic memory. The researchers trained the DRL agent to learn optimal waypoint generation in an end-to-end manner and validated their approach in 3D maps of typical residencies. 
Bansal et al. \cite{bansal2020combining} trained a learning-based perception module to set waypoints in unknown environments for a collision-free path. The model is trained using visual representations and is combined with conventional model-based control for navigation. In this work, we follow a similar hierarchical approach, but focus on employing a DRL-based local planner for the obstacle avoidance problem.
Gundelring et al. \cite{guldenringlearning} first integrated a DRL-based local planner with a conventional global planner from the ubiquitously used robot operating system (ROS) and demonstrated promising results. The researchers employs a subsampling of the global path to create waypoints for the DRL-local planner.
Similarly, Regler et al. \cite{regier2020deep} proposes a hand-designed sub-sampling to deploy a DRL-based local planner with conventional navigation stacks. 
A limitation of these works is that the a simple sub-sampling of the global path is inflexible and could lead to hindrance in complex situations, e.g. when multiple humans are blocking the way. 
In this work, we introduce an intermediate planner to spawn way-points more flexibly by considering local as well as global map information for optimal waypoint generation and replanning functionality. This should result in smoother trajectories, increased efficiency, and less collisions.

\section{Methodology}
In this chapter, we present the methodology of our proposed framework. In order to accomplish long-range navigation while utilizing a DRL-based local planner, our work proposes a combination with traditional global planners. The system design is described in Fig. \ref{intro}. We propose an interconnection entity called the intermediate planner, which connects the two planners and present two different approaches to generate waypoints. As a baseline approach, we also provide a simple subsampling of waypoints along the path as proposed by \cite{guldenringlearning} and \cite{regier2020deep}. This approach is denoted as SUB-WP.

\subsection{Spatial Horizon Waypoint Generator}
\begin{figure}[!h]
\centering
\includegraphics[width=0.2\textwidth]{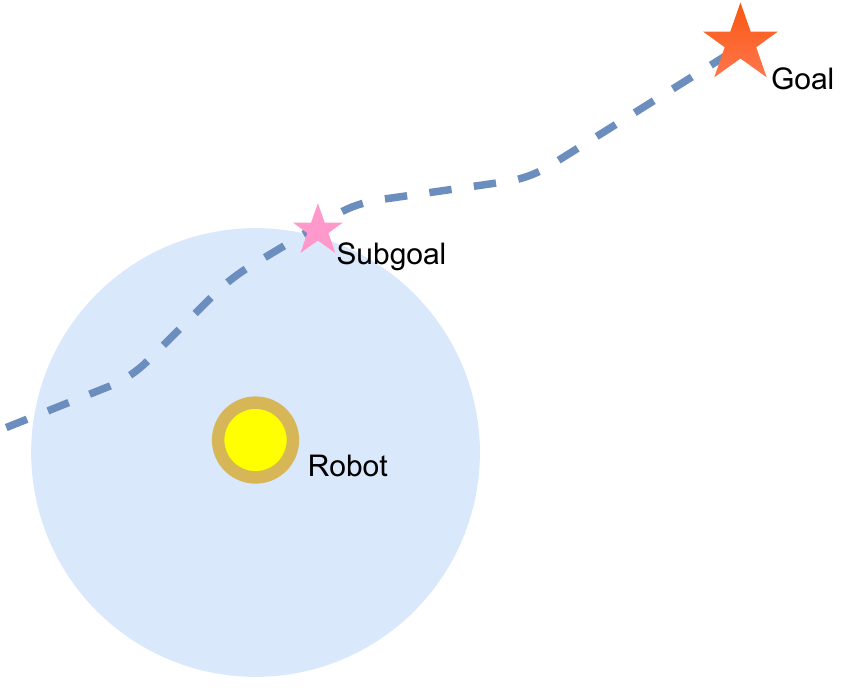}
\caption{Subgoal selection. Given a global path, our intermediate planner will calculate a subgoal based on the spatial horizon $dist_{ahead}$.}
\label{fig:subgoal0}
\end{figure}

Our second waypoint generator calculates the waypoint using a look-ahead distance and the robot's spatial position. Given the global path $\pi_g$ as an array Y consisting of $N_t$ poses, we sample a subset $Y^{\text{subgoals}}_i \subset X^{poses}= \{x_i, \dots, x_N\}$ of local goals based on the robot's position $p_r$ and a look-ahead distance $d_{ahead}$. The concept is illustrated in Fig. \ref{fig:subgoal0}. The algorithm of the S-WP is described in Alg. \ref{subsampling}.

\begin{algorithm}[!h]
  \SetAlgoLined
  \KwIn{ Global path $\pi_{g}$, Robot position $p_r$}
  \KwOut{Subgoal $g_{sub}$ }
  Parameter: $d_{ahead}$\;
  Find the set of intersection points of $R(p_r,d_{ahead})$ and global path $\pi_{g}$ as $\Phi$\;
  \eIf{$\Phi$ is not empty}
  {select the intersection point near to the global goal as $g_{sub}$}
  {
    call GLOBAL REPLAN $\rightarrow$ new global path $\pi_{g}^{new}$\;
    Subgoal calculation$(\pi_{g}^{new},p_r)$\;
  }
  \caption{Subgoal calculation}
  \label{subsampling}
\end{algorithm}

This subset of goals can be observed as local sub-goals $x_i = p_g$ and given as input for the DRL-based local planner. Additionally, we integrate a replanning mechanism to rework the global path when the robot is off-course for a certain distance from the global path, or if a time limit $t_{lim}$ is reached without movement, e.g. when the robot is stuck. In that case, a global replanning will be triggered and a new sub-goal will be calculated based on the new global plan and current robot position. This approach is denoted as spatial-time-horizon waypoint generator (STH-WP).

\subsection{Landmark-based Waypoint Generator}
Utilizing a simple subsampling approach can result in inefficient trajectories especially in highly dynamic environments, because the initially planned trajectory of the robot is constantly being changed when avoiding unknown obstacles. This may cause the robot to ignore more suitable paths in order to reach the static preset waypoints which might cause the trajectory to be less smooth and efficient. 
For our final waypoint generator, we postulate the assumption that not all points on the global path $\pi_g$ are equally important and that only critical waypoints have to be visited by the robot. These points are called landmarks $L$ in this paper. 
Our waypoint generator utilizes these specific landmarks to calculate optimal waypoint positions for the DRL local planner. This enables smoother trajectories and results in more efficient navigation. In the following, the landmark selection is described.

\subsubsection{Landmark Selection}
\begin{figure}[h]
    \centering
    \includegraphics[width=0.45\textwidth]{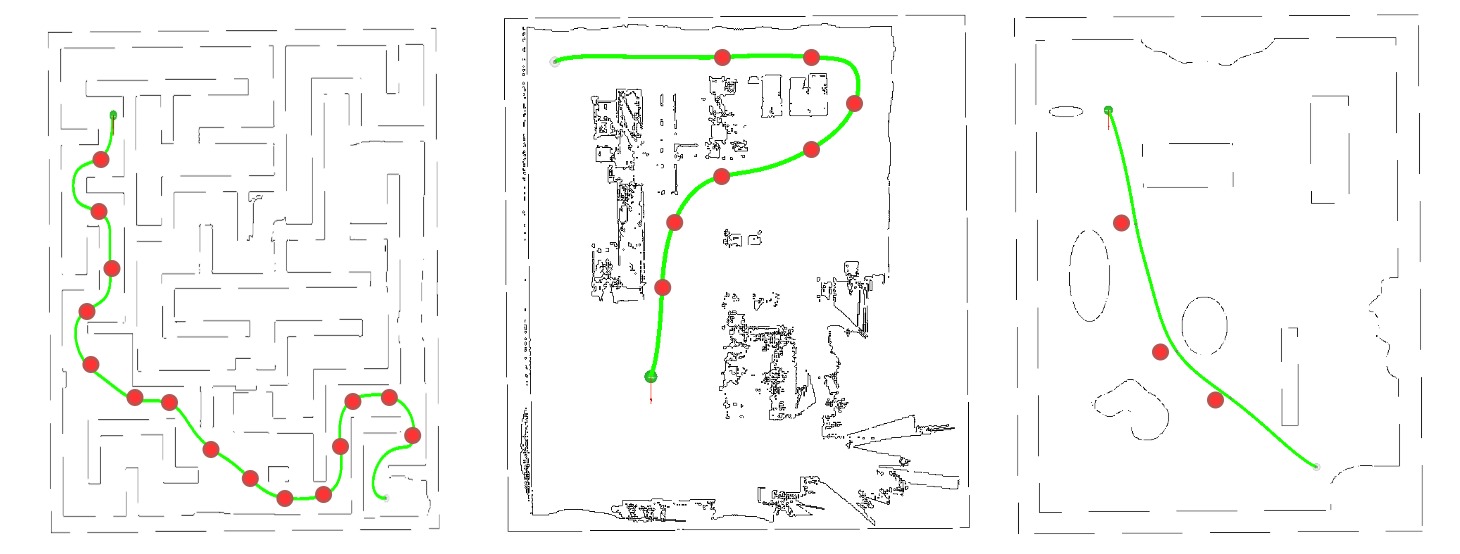}
    \caption{Landmark on different test maps.}
    \label{fig:landmark}
\end{figure}
In a map, landmarks are recognized by humans as turning points. In order to navigate across these points, a turning action is performed by the robot. In return, we formulate the landmark as a point where the robot's steering angle is bigger than a threshold value:
\begin{align}
    \psi_{p}>\psi_{thresh} \quad, \quad  p \in map 
    \label{equ:psi}
\end{align}
In order to calculate a reasonable steering angle value at a certain point on the map, the initial global path and its corresponding interpolated b-spline trajectory are used.
We use a kinodynamic A-star search global path planner method proposed by Zhou et al. \cite{kinoastar}, which provides b-spline properties to calculate reasonable steering angles for certain positions on the map.

\begin{algorithm}[!h]
  \SetAlgoLined
  \KwIn{ start position $p_{s}$, goal position $p_g$, occupancy gridmap}
  \KwOut{queue of landmarks $L$ }
  Parameter: $\psi_{thresh}$\;
  Global path $\pi_{g}$ $\leftarrow$ kinodynamicAstar ($p_{s}$,$p_g$)\;
  Parameterize global path $\pi_{g}$ as uniform b-spline trajectory $\pi^{*}_{g}$\;
  \For{$t$ from $t_{q}$ to $t_{{M}-{q}}$}{
    Evaluate position $p_t$, velocity  $v_t$, acceleration $a_t$ on  $\pi^{*}_{g}$  at time $t$\ using De Boor's algorithm\;
    Calculate steering angle velocity $\omega_{t}$ and steering angle $\psi_{t}$\;
    \If{$\psi_{t}>\psi_{thresh}$}{
        $L$.pushback($p_t$)\;
    }
  }
  $L$.pushback($p_g$)\;
  \caption{Landmark selection}
  \label{subsampling}
\end{algorithm}

Once the global b-spline trajectory is parameterized, the position, velocity, and acceleration at each time  $t \in [t_{q}, t_{{M}-{q}}] $ is be calculated. To evaluate the position at time $t$, we employ De Boor's algorithm \cite{deboor}. Finally, the derivatives of a point on the b-spline trajectory are calculated with  \cite{kinoastar}

\begin{align}
    \mathbf{v}_i=\frac{1}{\Delta{t}}(\mathbf{p}_{i+1}- \mathbf{p}_{i}) \quad 
    \mathbf{a}_i=\frac{1}{\Delta{t}}(\mathbf{v}_{i+1}- \mathbf{v}_{i})
    \label{equ:derivative}
\end{align}

The steering angle is calculated using the kinematic equations in Eq.\ref{equ:steering1}-\ref{equ:steering3}

\begin{align}
    \mathbf{a}_{norm}= \frac{\mathbf{v} \times
    \mathbf{a}}{\|\mathbf{v}\|} \label{equ:steering1}\\
    \omega= \frac{\|\mathbf{a}_{norm}\|}{\|\mathbf{v}\|} \label{equ:steering2}\\
    \psi =\int{\omega}dt
    \label{equ:steering3}
\end{align}
Fig. \ref{fig:landmark} illustrates landmarks on different maps using aforementioned calculations.

\subsubsection{Subgoal Calculation}
\begin{figure}
    \centering
    \includegraphics[width=0.27\textwidth]{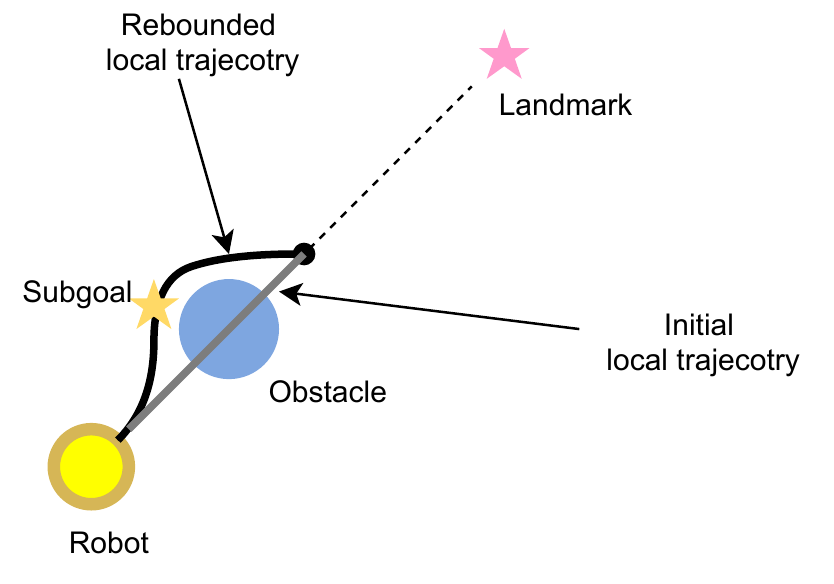}
    \caption{Subgoal calculation for the LM-WP approach.}
    \label{fig:subgoal}
\end{figure}

\begin{figure*}[!h]
   
\begin{subfigure}{0.29\textwidth}(a)
 \includegraphics[width=\linewidth,height=2in]{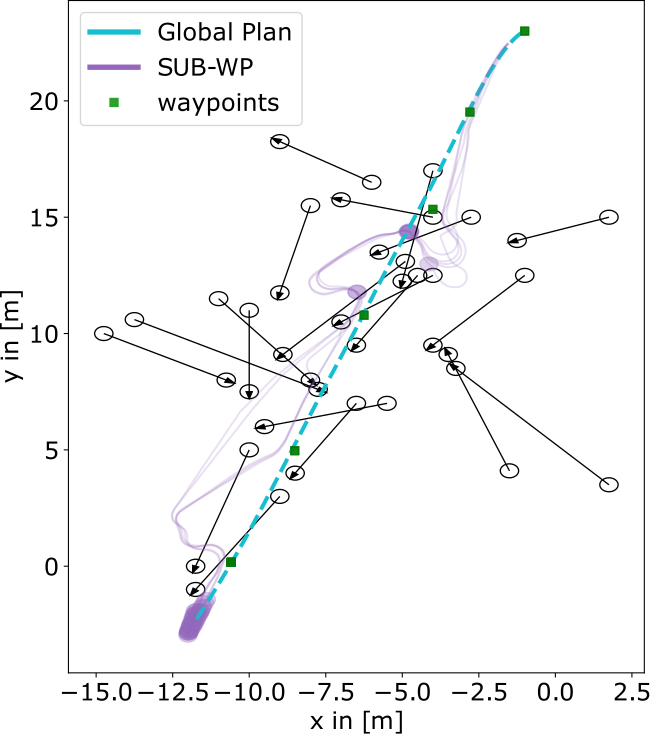}

  \label{fig:1}
\end{subfigure}\hfil 
\begin{subfigure}{0.29\textwidth }(b)
\includegraphics[width=\linewidth,height=2in]{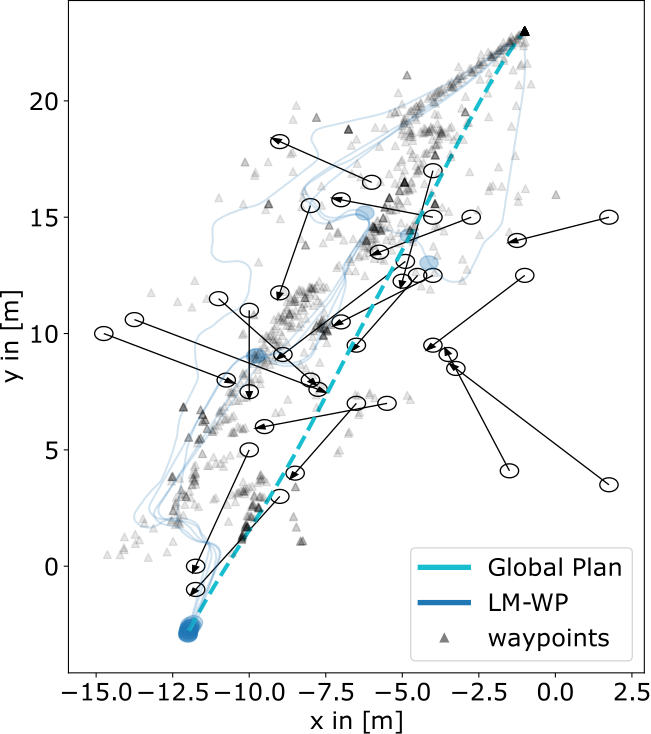}

  \label{fig:2}
\end{subfigure}\hfil 
\begin{subfigure}{0.29\textwidth}(c)
\includegraphics[width=\linewidth,height=2in]{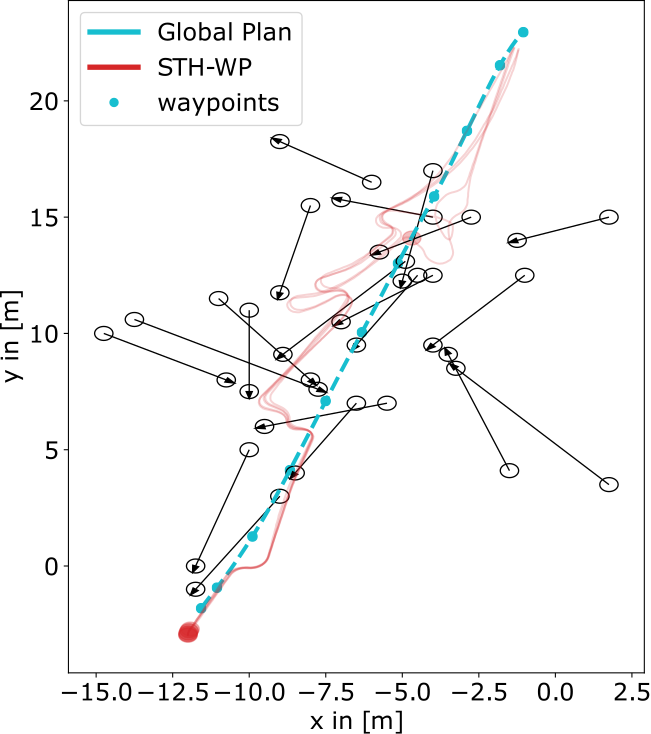}

  \label{fig:3}
\end{subfigure}

\medskip
\begin{subfigure}{0.29\textwidth}(d)
\includegraphics[width=\linewidth,height=2in]{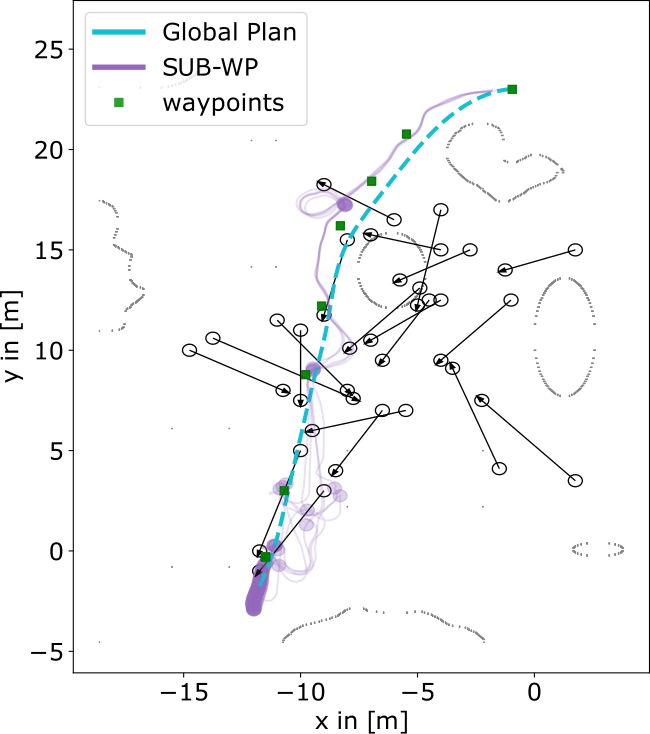}

  \label{fig:4}
\end{subfigure}\hfil 
\begin{subfigure}{0.29\textwidth}(e)
 \includegraphics[width=\linewidth,height=2in]{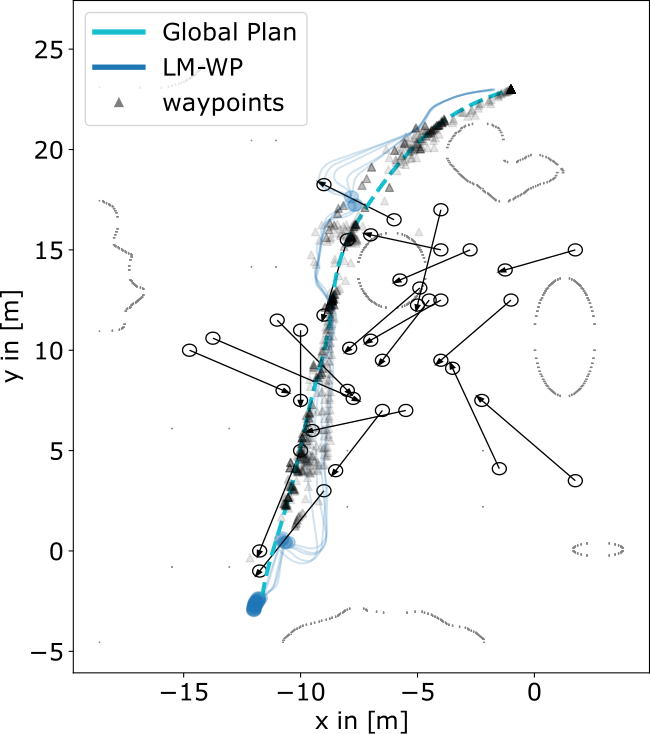}

  \label{fig:5}
\end{subfigure}\hfil 
\begin{subfigure}{0.29\textwidth}(f)
 \includegraphics[width=\linewidth,height=2in]{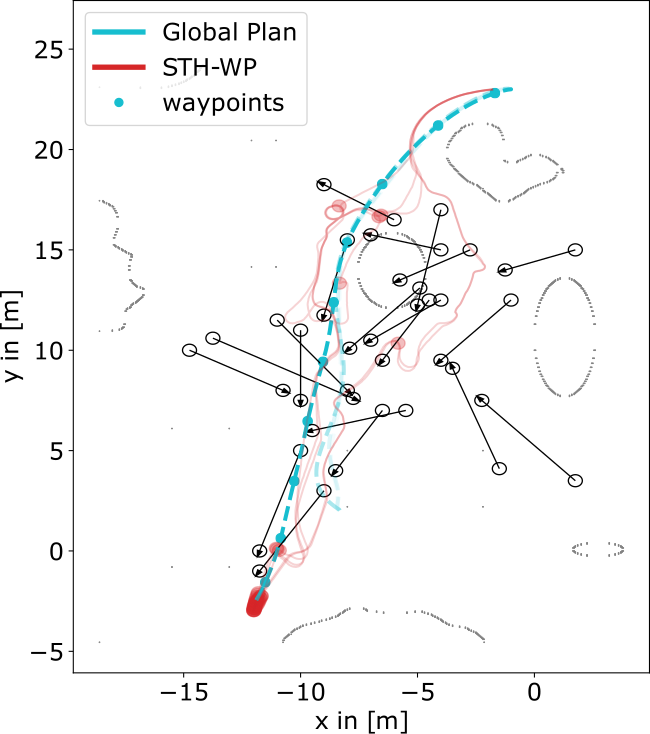}

  \label{qual}
\end{subfigure}
\caption{Trajectories and waypoints of all planners on the test scenario with 20 obstacles and a obstacle velocity of $v_{obs}=0.3 m/s$. Upper row: plain test area, lower row: office map.  (a) SUB-WP on the empty map (b) STH-WP on the empty map (c) LM-WP on the empty map (d) SUB-WP on the office map (e) STH-WP on the office map (f) LM-WP on the office map}
\label{quali1122}
\end{figure*}

\begin{algorithm}[]
  \SetAlgoLined
  \KwIn{ current position $p_{t}$, landmark position $p_{l}$, occupancy gridmap}
  \KwOut{subgoal $g_{sub}$ }
  Parameter: $d_{ahead}$, $t_{eval}$\;
  1. Calculate a straight one-shot initial local trajectory $\zeta_{0}$ from $p_{t}$ towards $p_{l}$ with  length $d_{ahead}$\;
  2. Optimize the initial trajectory $\zeta_{0}$ using ESDF-free gradient-based methods to get a collision-free rebounded b-spline trajectory $\zeta_{t}$\;
  3. Select the subgoal $g_{sub}$ on $\zeta_{t}$ at time $t_{eval}$ using  De Boor's algorithm\;
  \caption{Subgoal calculation}
  \label{subsampling}
\end{algorithm}

Once landmarks are selected from the global path, the waypoints will be calculated according to the robot's current location, the current approaching landmark location, and the obstacles' locations in the occupancy grid-map. Hence, the calculated sub-goal is not only calculated with respect to the landmark but also takes obstacles within sensor range into consideration. To ensure a collision-free path, we optimize the b-spline trajectory using the ESDF-free gradient-based optimization method introduced by Zhou et al. \cite{egoplanner}. This method has the advantage of fast calculation times, which makes online deployment feasible. After the local trajectory optimization is finised, the sub-goal is calculated by evaluating a point on the optimized b-spline trajectory in time $t_{eval}$ with help of De Boor's algorithm. The algorithm is described in Alg. \ref{subsampling}.

\begin{figure*}[!h]
\centering
\begin{subfigure}[b]{0.87\textwidth}
   \includegraphics[width=1\linewidth]{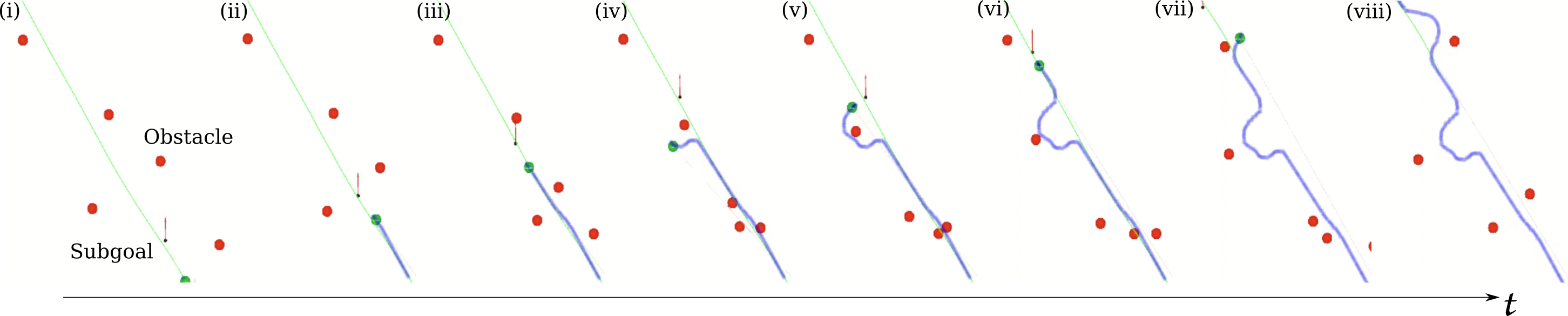}
   \caption{SUB-WP}
   \label{fig:Ng2}
      \bigskip
\end{subfigure}

\begin{subfigure}[b]{0.87\textwidth}
   \includegraphics[width=1\linewidth]{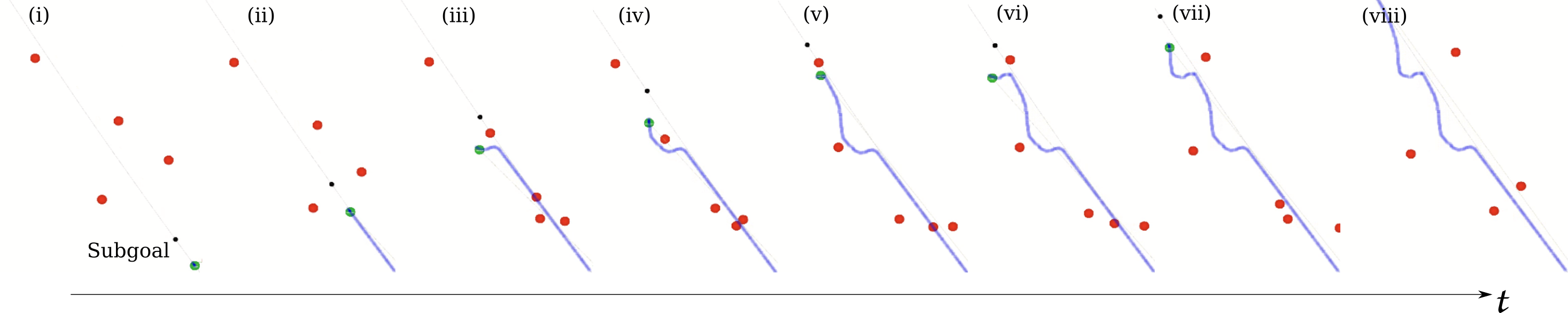}
   \caption{STH-WP}
   \label{fig:Ng2}
      \bigskip
\end{subfigure}

\begin{subfigure}[b]{0.87\textwidth}
   \includegraphics[width=1\linewidth]{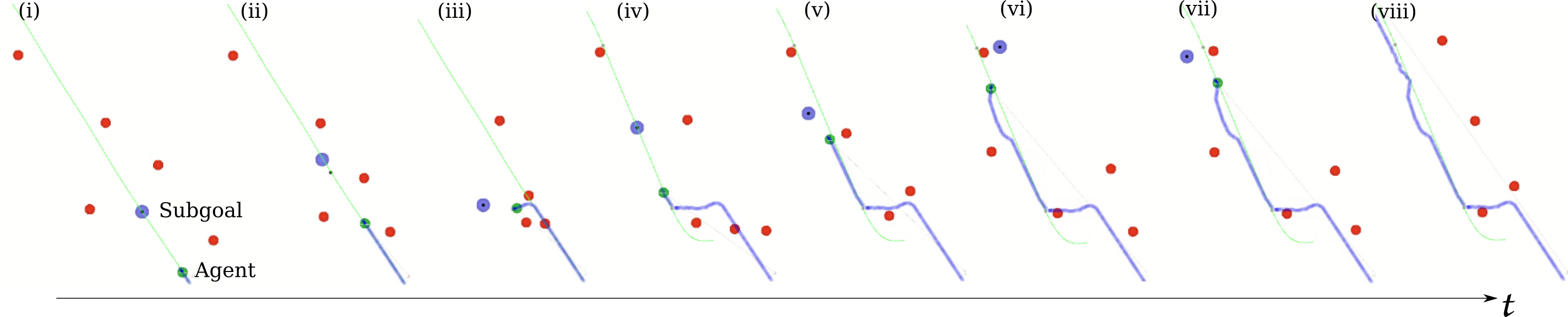}
   \caption{LM-WP}
   \label{fig:Ng1}
\end{subfigure}

\caption{Comparison of different waypoint generators. The trajectories are plotted over time. It is observable that LM-WP dynamically generates waypoints based on the obstacle position and results in a shorter, smoother trajectory (s. (a) (vii)). The path for SUB-WP is the most inefficient and includes roundabouts circles around obstacle areas. STH-WP achieves an efficient navigation as well without replanning and can follow the global path in a more robust manner. The sequences and other situations can be observed in action in the supplementary video material.}
\label{timeplot}
\end{figure*}

\section{Results and Evaluation}
We conducted experiments on two different maps in which we created a total of six different test scenarios of increasing difficulty with five, 10 and 20 obstacles. In each scenario, the obstacle velocities are set to 0.3 m/s. More extensive evaluations of the obstacle velocities' impacts are presented in the quantitative evaluations. As a global planner, the optimized b-spline kino A-star planner is used for all approaches. Localization is assumed to be perfectly known. For each planner, we conduct 30 test runs on each scenario. 
Finally, we compare our joint systems against conventional ones- TEB \cite{rosmann2015timed} and MPC \cite{rosmann2019time}- in terms of safety, robustness and efficiency. 
For STH-WP, the time and spatial horizon is set to $t_{lim}=4 s$ and $d_{ahead}=1.55 m$ respectively. The lookahead distance of the SUB-WP approach is set to be 1 m.

\subsection{Evaluation of Waypoint Generators}
First, we evaluate the overall performance of our waypoint generators against one another. Figure \ref{quali1122} illustrates the qualitative trajectories and waypoints of each planner. The opacity of the waypoints indicates their order of generation.
It is observed that the trajectories of the SUB-WP contain more circular movements and are less efficient compared to the other two approaches. Due to the preset waypoints of the SUB-WP approach, the agent has to strictly reach each waypoint regardless of a potentially more efficient trajectory. This happens once the agent is off of the path when avoiding obstacles, which makes the approach inflexible and results in less efficient trajectories. As expected, the paths of LM-WP have the most variance due to the constant replanning as obstacles are being observed. Most of the waypoints are placed on the global path. However, in situations where obstacles are interfering, the waypoints are adjusted dynamically as can be observed in Fig. \ref{quali1122} (b), (e). This results in a more compact path with less collisions.
It is notable that whereas the LM-WP achieves robust navigation with efficient paths on the office map, in the empty map, it is less efficient and performs more roundabout paths. This is caused by the lack of landmarks as reference points for the intermediate planner to calculate optimal waypoints, which indicates that our approach works better in structured environments. Contrarily, the STH-WP achieves robust and strict trajectories towards the goal on the empty map. Whereas in the office map, the more roundabout paths are observed. Nevertheless, in all scenarios, the paths are still more efficient compared to the SUB-WP- which has increased circular movements. This is more evident in the office map (Fig. \ref{quali1122} (d)).
In Fig. \ref{timeplot}, the trajectories of one test run in a scenario with five obstacles were plotted over time. Once again, it is evident that the SUB-WP method results in a more inefficient trajectory because the robot has to follow the preset waypoint while dismissing potentially more suitable trajectories. The dynamic adjustment of the LM-WP generator results in a shorter and smoother path. Similarly, the STH-WP approach results in smoother trajectory because the waypoint adapts to the robots position. Live demonstrations of the approaches can be viewed in the supplementary video material.
\begin{figure*}[!h]
   
\begin{subfigure}{0.29\textwidth}(a)
\includegraphics[width=\linewidth,height=2in]{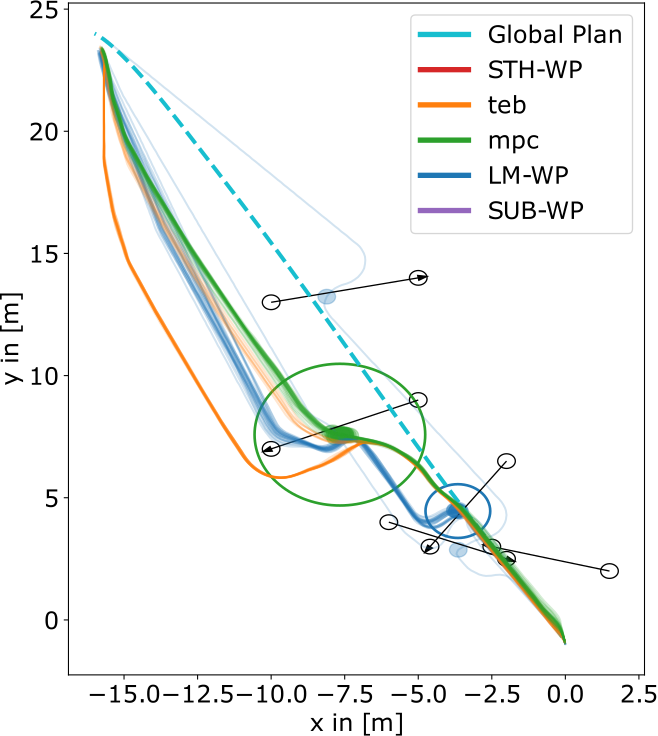}

  \label{fig:1}
\end{subfigure}\hfil 
\begin{subfigure}{0.29\textwidth }(b)
  \includegraphics[width=\linewidth,height=2in]{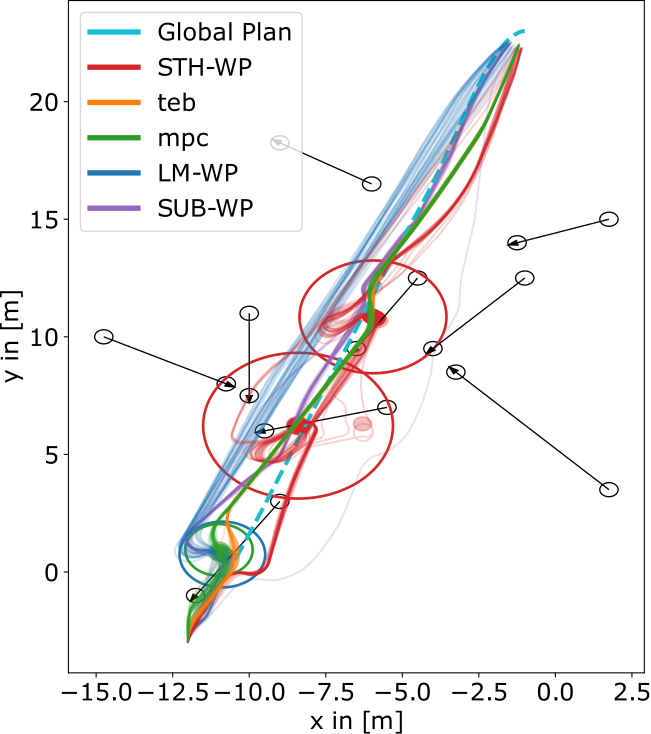}

  \label{fig:2}
\end{subfigure}\hfil 
\begin{subfigure}{0.29\textwidth}(c)
 \includegraphics[width=\linewidth,height=2in]{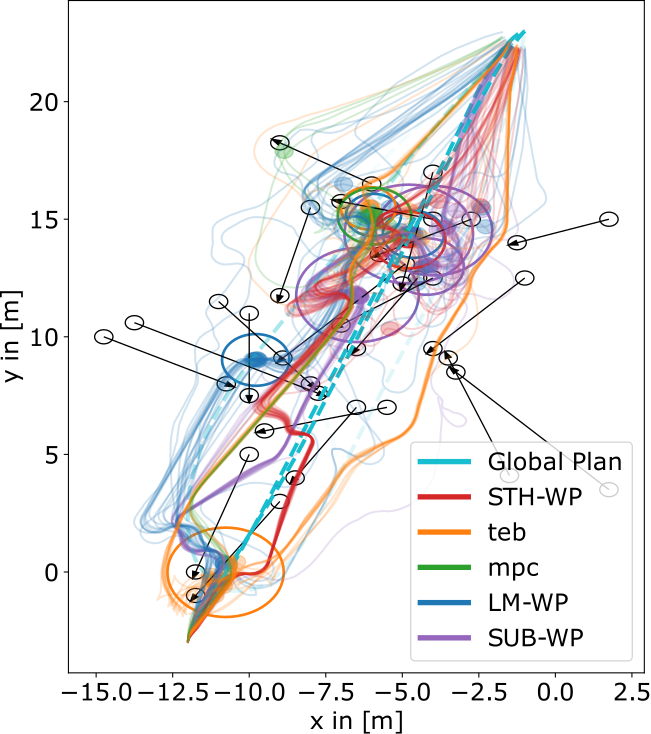}

  \label{fig:3}
\end{subfigure}

\medskip
\begin{subfigure}{0.29\textwidth}(d)
 \includegraphics[width=\linewidth,height=2in]{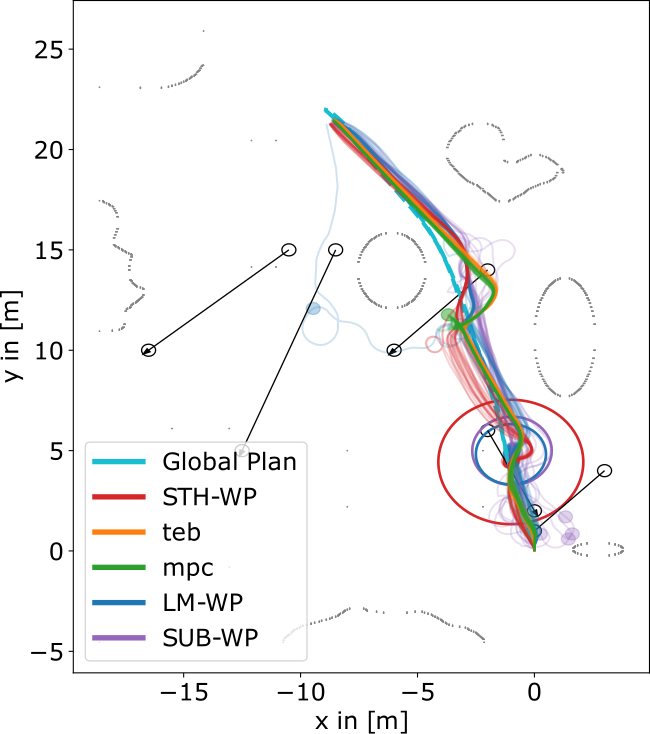}

  \label{fig:4}
\end{subfigure}\hfil 
\begin{subfigure}{0.29\textwidth}(e)
 \includegraphics[width=\linewidth,height=2in]{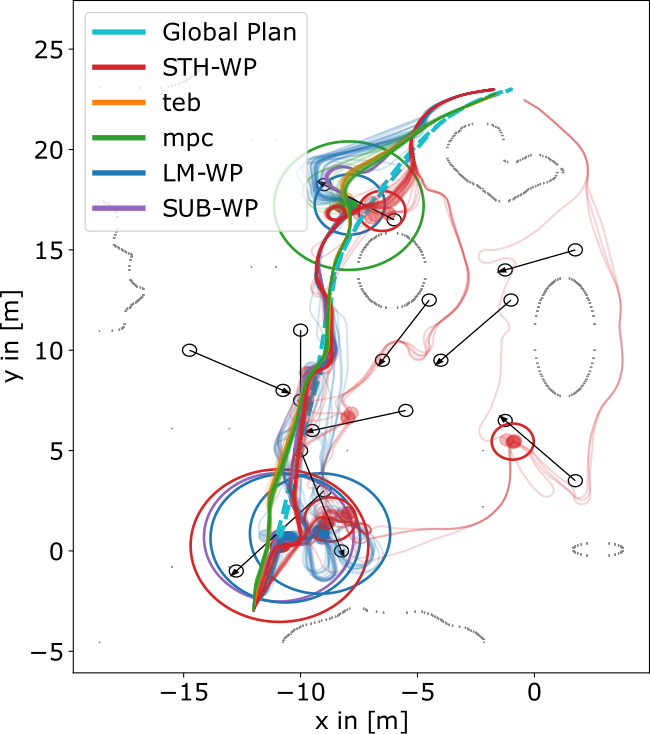}

  \label{fig:5}
\end{subfigure}\hfil 
\begin{subfigure}{0.29\textwidth}(f)
 \includegraphics[width=\linewidth,height=2in]{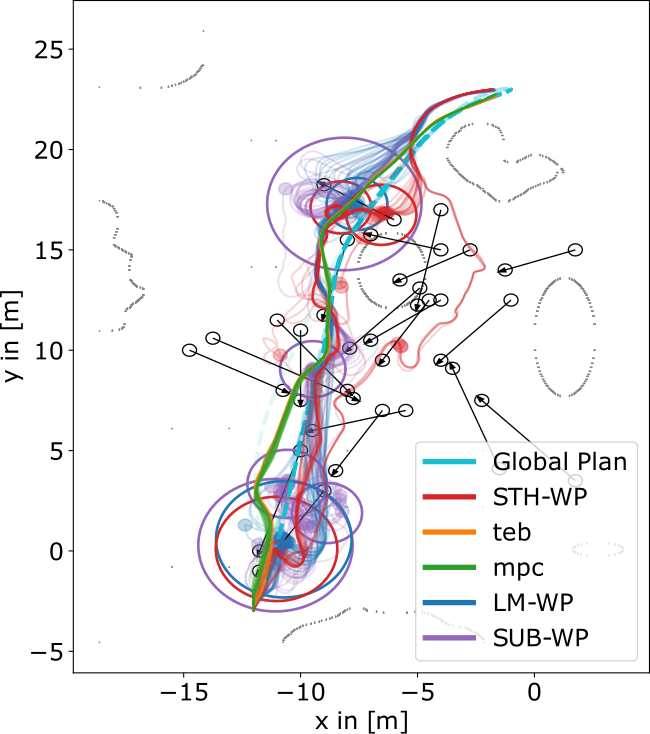}

  \label{quali3333}
\end{subfigure}
\caption{Trajectories of all planners on different test scenarios with obstacle velocity $v_{obs}=0.3 m/s$ and 20 obstacles. Upper row: empty map, lower row: office map. (a) SUB-WP on the empty map (b) STH-WP on the empty map (c) LM-WP on the empty map (d) SUB-WP on the office map (e) STH-WP on the office map (f) LM-WP on the office map. The obstacles trajectories are visualized in black. Collisions are visualized as dots in the respective colors. The circle radius' denotes the number of collisions (a larger radius means more collisions).}
\label{quali11}
\end{figure*}

\subsection{Comparison with Conventional Navigation Systems}
After evaluating our approaches against one another, we compare the navigation behavior of our joint navigation systems against model-based navigation systems MPC \cite{rosmann2019time} and TEB \cite{rosmann2015timed}.
Our joint systems includes the following: our proposed waypoint generators as intermediate planner, the kino A-star global planner, and a DRL-based local planner of our previous work \cite{kastner2020deep} against the model-based navigation systems MPC \cite{rosmann2019time} and TEB \cite{rosmann2015timed}. Throughout this paper, the notion previously used to denote the waypoint generators are used to denote the joint systems. The traditional navigation systems, consisting of the same kino A-star global planner and the model-based local planners MPC and TEB, are denoted as MPC and TEB respectively. The qualitative trajectories of all systems for an obstacle velocity of 0.3 m/s on all scenarios are illustrated in Fig. \ref{quali11}. The intensity of the trajectories indicates their frequency. Results for all test runs are stated in Table \ref{quanttable}. Additionally, we plotted collision circles, which radius' indicates the number of collisions.

\subsubsection{Safety and Robustness}
\begin{figure*}[!h]
    \centering 
    \subcaption{\textbf{Empty Map}}
\begin{subfigure}{0.22\textwidth}(i)
 \includegraphics[width=\linewidth]{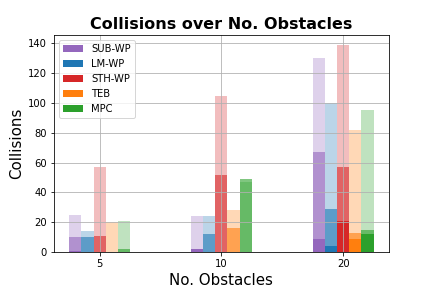}

  \label{fig:1}
\end{subfigure}\hfil 
\begin{subfigure}{0.22\textwidth}(ii)
 \includegraphics[width=\linewidth]{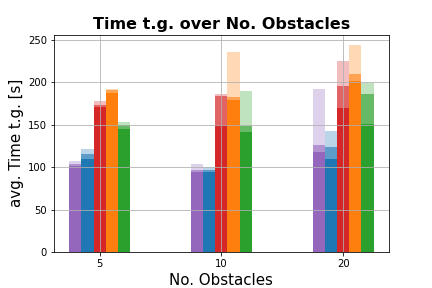}

  \label{fig:2}
\end{subfigure}\hfil 
\begin{subfigure}{0.22\textwidth}(iii)
 \includegraphics[width=\linewidth]{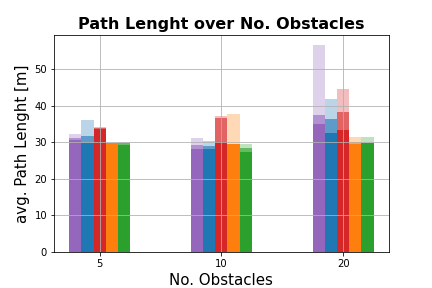}

  \label{fig:3}
\end{subfigure}
\begin{subfigure}{0.22\textwidth}(iv)
\includegraphics[width=\linewidth]{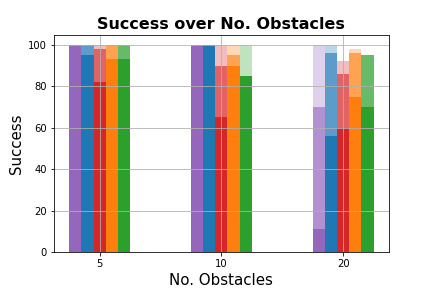}

  \label{fig:3}
\end{subfigure}
\medskip
\begin{subfigure}{0.22\textwidth}(v)
\includegraphics[width=\linewidth]{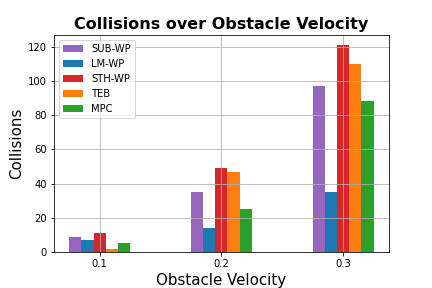}

  \label{fig:4}
\end{subfigure}\hfil 
\begin{subfigure}{0.22\textwidth}(vi)
 \includegraphics[width=\linewidth]{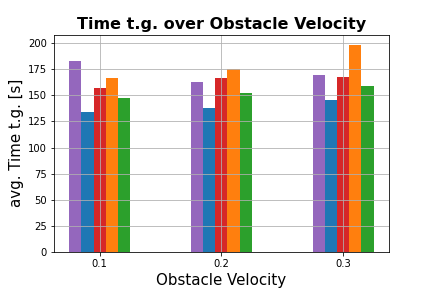}

  \label{fig:5}
\end{subfigure}\hfil 
\begin{subfigure}{0.22\textwidth}(vii)
\includegraphics[width=\linewidth]{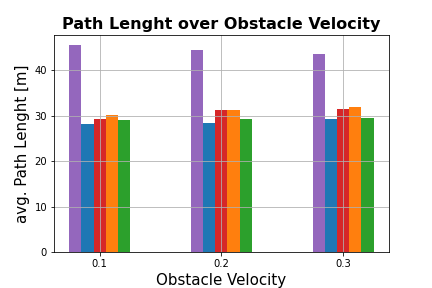}

  \label{fig:6}
\end{subfigure}
\begin{subfigure}{0.22\textwidth}(viii)
\includegraphics[width=\linewidth]{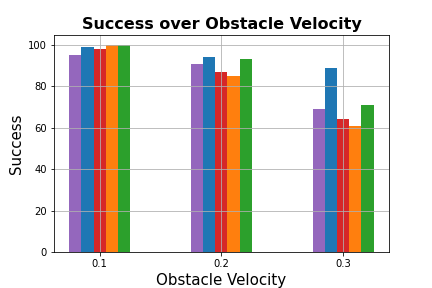}

  \label{fig:3}
\end{subfigure}
\medskip
\subcaption{\textbf{Office Map}}
\begin{subfigure}{0.22\textwidth}(i)
\includegraphics[width=\linewidth]{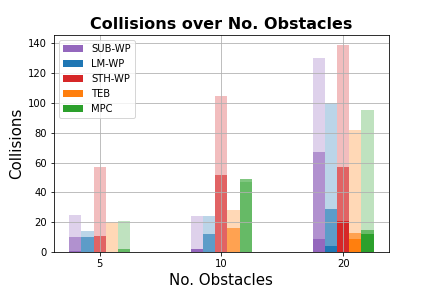}

  \label{fig:4}
\end{subfigure}\hfil 
\begin{subfigure}{0.22\textwidth}(ii)
 \includegraphics[width=\linewidth]{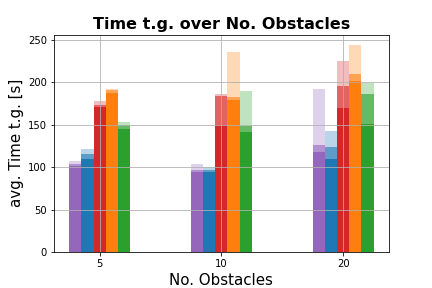}

  \label{fig:5}
\end{subfigure}\hfil 
\begin{subfigure}{0.22\textwidth}(iii)
\includegraphics[width=\linewidth]{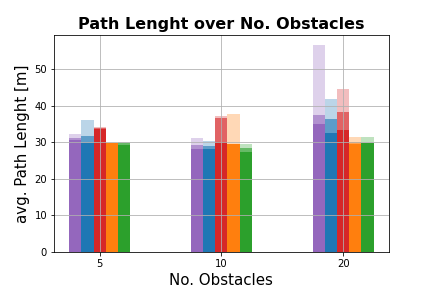}

  \label{fig:6}
\end{subfigure}
\begin{subfigure}{0.22\textwidth}(iv)
 \includegraphics[width=\linewidth]{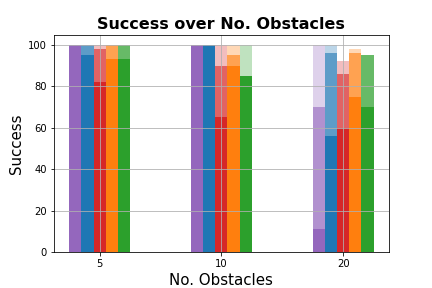}

  \label{fig:3}
\end{subfigure}

\medskip
\begin{subfigure}{0.22\textwidth}(v)
  \includegraphics[width=\linewidth]{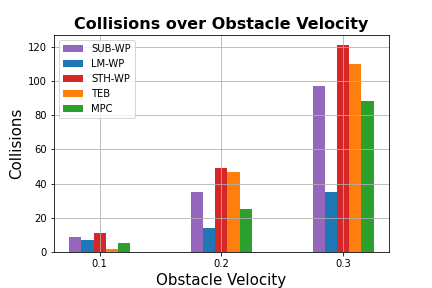}

  \label{fig:4}
\end{subfigure}\hfil 
\begin{subfigure}{0.22\textwidth}(vi)
\includegraphics[width=\linewidth]{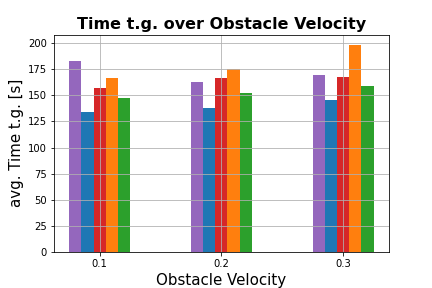}

  \label{fig:5}
\end{subfigure}\hfil 
\begin{subfigure}{0.22\textwidth}(vii)
\includegraphics[width=\linewidth]{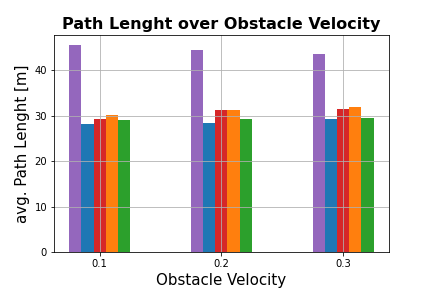}

  \label{fig:6}
\end{subfigure}
\begin{subfigure}{0.22\textwidth}(viii)
 \includegraphics[width=\linewidth]{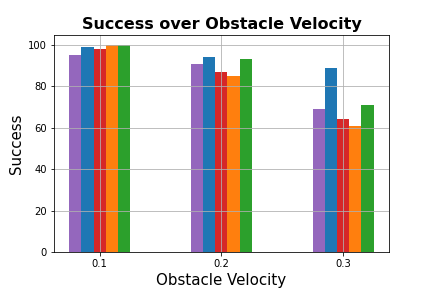}

  \label{fig:3}
\end{subfigure}
\caption{Results from test-runs on (a) the empty map and (b) office map. (i) Collisions over number of obstacles and $v_{obs}$, (ii) average time to reach goal over number of obstacles and $v_{obs}$, (iii) Average path length over number of obstacles and $v_{obs}$. (iv) Success over number of obstacles and $v_{obs}$. With $v_{obs}=(0.1,0.2,0.3) m/s$, while a high opacity indicates a lower velocity. (v) Collisions over obstacle velocities, (vi) average time to reach goal over obstacle velocities (lower is better), (vii) Path length over obstacle velocity. (viii) Success rate over obstacle velocity}
\label{quantbig}
\end{figure*}

It is observed that all navigation systems manage a robust collision avoidance and efficient navigation towards the goal for the scenarios with five obstacles. With increasing number of obstacles, collisions increase which is evident in larger circles.
Less collisions and thus smaller circles are observed for systems with our proposed LM-WP and STH-WP generators compared to TEB. Furthermore, the paths of our joint systems are more efficient compared to the TEB planner, which performs more roundabout paths. The model-based MPC planner accomplishes a similar performance in terms of path length and navigation efficiency. However, with more obstacles, the collisions increased (s. Fig. \ref{quali11} (e)).
While SUB-WP manages to keep a competitive performance in maps with 5 obstacles, increased collisions occur in maps with 10 and 20 obstacles (s. \ref{quali11} (e),(f)). 
In terms of collision rates and robustness, the LM-WP approach was best performing due to its replaning capability and long sight, which calculates the next waypoints online. This way, even in stuck situations the planner manages to find a way out. In contrast, the SUB-WP method was stuck in hard situations when obstacles were constantly blocking the way. The STH-WP approach was limited in such situations as well.
The quantitative results are listed in Table \ref{quanttable} and plotted in Fig. \ref{quantbig}. We evaluated the time to reach the goal, the path length, the number of collisions and the success rate for all navigation systems. Thereby, runs with less than three collisions are regarded as successful.
Our LM-WP approach achieves the best overall performance in terms of collision and success rate accomplishing an average of 95.94\%. This holds true for scenarios with increased obstacle count and velocities.
Notably, the SUB-WP approach accomplishes a similar performance in maps with five obstacles but experiencing a rapid decline in maps with 10 and 20 obstacles and $v_{obs}=0.2$ and $0.3 m/s$ accomplishing only 40.5 and 16.5\% respectively. 
This decline in performance for scenarios with 10 and 20 obstacles is also evident for MPC and TEB achieving only over 68\% success compared to 72.5\% by our LM-WP.
\begin{table*}[!h]

	\centering
	\setlength{\tabcolsep}{3pt}
	\renewcommand{\arraystretch}{1}
	\caption{Quantitative evaluations}
	\begin{tabular}{p{2.2cm}|p{1cm}p{1cm}p{1cm}p{1cm}|p{1cm}p{1cm}p{1cm}p{1cm}|p{1cm}p{1cm}p{1cm}p{1cm}}
		\hline
		\hline
		     & Time [s]& Path [m]  & Collisions &  Success & Time [s]& Path  [m]  & Collisions &  Success & Time [s]& Path [m]  & Collisions &  Success \\ \hline

		\textbf{$v_{obs}$ 0.1m/s}& \multicolumn{4}{c}{\bfseries 5 dyn. Obstacles} & \multicolumn{4}{c}{\bfseries 10 dyn. Obstacles} & \multicolumn{4}{c}{\bfseries 20 dyn. Obstacles} \\ 
		\cline{1-13}
SUB-WP &  \textbf{100.95} &   29.10 &          22.5 &       96.5 &   \textbf{97.56} &   29.26 &           1.0 &      100.0&  150.71 &   40.30 &          59.0 &       57.5 \\
LM-WP              &   103.39 &   27.09 &           2.0 &      100.0 &   99.61 &   \textbf{27.71} &           1.0 &      100.0 &  \textbf{122.21} &   30.32 &           \textbf{5.5 }&       99.5 \\
STH-WP   &  158.69 &   31.08 &           0.0 &      100.0 &  151.22 &   29.98 &           1.0 &      100.0 &  163.50 &   31.31 &          16.0 &       95.0 \\
TEB               &  171.95 &   29.28 &           0.0 &      100.0 &  171.83 &   28.70 &           2.5 &      100.0 &  183.86 &   29.75 &           \textbf{5.5} &       99.0 \\
MPC               &  134.45 &   \textbf{26.95} &           0.0 &      100.0 &  138.16 &   26.28 &           0.0 &      100.0 &  149.25 &   \textbf{29.34} &           8.5 &       \textbf{97.5} \\
	    \cline{1-13} \hline
	    \textbf{$v_{obs}$ 0.2m/s} \\
	    \cline{1-13}
SUB-WP &   131.09 &   28.36 &          14.5 &      100.0 &  137.28 &   31.94 &           \textbf{6.0} &      100.0 &  144.68 &   40.94 &          87.5 &       40.5 \\
LM-WP              &   \textbf{99.95} &   28.40 &           7.0 &      100.0 &  \textbf{103.55} &   28.19 &           6.5 &      100.0 &  \textbf{130.71} &   32.45 &          21.5 &       \textbf{95.0} \\
STH-WP   &  1150.46 &   29.70 &          11.0 &       97.5 &  184.04 &   32.51 &          44.0 &       90.0 &  181.15 &   34.72 &          53.0 &       86.5 \\
TEB               &  179.35 &   29.44 &           4.5 &       98.0 &  176.62 &   29.20 &          18.0 &       94.5 &  192.76 &   30.62 &          30.0 &       90.5 \\
MPC               &  147.97 &   \textbf{28.32} &           \textbf{2.0} &      \textbf{100.0} &  144.76 &   \textbf{27.18} &          28.5 &       91.0 &  169.28 &   \textbf{29.60} &          \textbf{20.0} &       94.0 \\
	    \cline{1-13} \hline
	    \textbf{$v_{obs}$ 0.3m/s} \\
	    \cline{1-13}
SUB-WP &  106.83 &   30.33 &          26.5 &       \textbf{98.0} &  130.81 &   37.93 &          66.0 &       57.5 &  170.80 &   50.11 &         129.5 &       16.5 \\
LM-WP              &   \textbf{101.56} &   30.16 &          \textbf{11.5} &       97.5 &  \textbf{106.53} &   29.35 &          \textbf{15.0} &       \textbf{99.0} & \textbf{ 144.07} &   35.57 &          \textbf{67.5} &      \textbf{ 72.5} \\
STH-WP   &  153.38 &   29.96 &          43.5 &       86.5 &  185.96 &   37.17 &          73.5 &       75.0 &  206.20 &   40.02 &         130.0 &       62.0 \\
TEB               &  185.80 &   \textbf{28.96} &          23.5 &       91.0 &  212.54 &   33.43 &          44.0 &       85.5 &  221.21 &   31.65 &          96.0 &       68.0 \\
MPC               &  149.05 &   29.95 &          24.5 &       92.0 &  169.80 &   \textbf{27.90} &          \textbf{40.0} &       88.0 &  178.74 &   \textbf{30.48} &          \textbf{91.5} &       \textbf{70.5} \\
	    \cline{1-13}
		\hline
		\hline
	    & \multicolumn{4}{c}{\textbf{Overall Average}} & \multicolumn{4}{c}{\bfseries Empty Map} & \multicolumn{4}{c}{\bfseries Office Map} \\ 
	    \cline{1-13}
	    SUB-WP &   142.74 &     35.03 &          40.83 &       78.22 &      116.14 &    34.57 &          29.78 &       86.78 &   169.35 &    35.50 &          51.89 &       69.67 \\
LM-WP              &    \textbf{112.40} &     29.92 &          \textbf{15.28} &       \textbf{95.94} &      \textbf{112.76} &    32.87 &          21.44 &       94.11 &   \textbf{112.04} &    \textbf{26.96} &           \textbf{9.11} &      \textbf{ 97.78 }\\
STH-WP   &   169.40 &     32.38 &          41.33 &       88.06 &      181.34 &    35.73 &          49.11 &       85.89 &   157.46 &    29.03 &          33.56 &       90.22 \\
TEB               &  188.44 &     30.12 &          24.89 &       91.83 &      202.75 &    30.66 &          \textbf{18.67} &       94.11 &   174.12 &    29.57 &          31.11 &       89.56 \\
MPC               &    153.50 &     \textbf{28.44} &          23.89 &       92.56 &      162.70 &    \textbf{29.51} &          26.78 &       91.33 &   144.29 &    27.38 &          21.00 &       93.78 \\
	    \cline{1-13}
	   \cline{1-13}
		\hline
		\hline
		
	\end{tabular}

	\label{quanttable}
	
\end{table*}

\subsubsection{Efficiency}
In terms of efficiency, LM-WP achieves an overall average of 112.4 s as opposed to more then 140 meters by the other approaches. The MPC system requires a smaller path of 28.44 m while requiring more time (avg. of 153.50 m). Furthermore, MPC is slightly less efficient with 150.2 m and 29.2 seconds. Similar to the success and collision rates, a decline in efficiency for all model-based planners can be noticed for scenarios with increased obstacle numbers and velocities. As stated in Table \ref{quanttable}, TEB and MPC require an average of 188.43 and 153.49 s respectively with an average path length of 30.11 and 28.44 meters respectively, whereas LM-WP, SUB-WP require 112.4 s and 142.74 s while the average path length is 29.92 and 35.03 meters respectively. 
It is notable that although the TEB system accomplishes a short average path length, it took the longest time to reach the goal. This is because TEB tends to stop and wait once obstacles are approaching. Overall, our proposed joint systems using LM-WP and STH-WP achieve shorter trajectories and less collisions in scenarios with more than 10 obstacles and velocities higher than 0.2 m/s. Due to the DRL local planner of our approaches, obstacles can be avoided more efficiently.

\section{Conclusion}
In this paper, we proposed waypoint generators to connect DRL-based local planners with conventional global planners for long-range-navigation. Therefore, we propose two different methods: a spatial and time horizon based waypoint generator which calculates the waypoint based on the robots state and a build in replanning mechanism, and a landmark based waypoint generator, which generates waypoints while taking obstacle and robot positions and global information into consideration. We integrated our approaches into the ubiquitously used ROS navigation stack and combined traditional global planner with a DRL-based local planner from our previous work. We demonstrated the feasibility of our approach and compared it against model-based navigation systems. Results showed that our joint navigation systems outperform conventional systems in terms of safety and efficiency. Our ongoing work focuses on extensive evaluations in more complex maps like mazes or office buildings. Furthermore, we aspire to deploy a DRL-based waypoint generator which learns to spawn optimal waypoints using sensor observations.

\appendix
The code is publicly available at https://github.com/ignc-research/arena-rosnav.




\addtolength{\textheight}{-1cm} 




\typeout{}
\bibliographystyle{IEEEtran}
\bibliography{ref}

\begin{thebibliography}{10}
\providecommand{\url}[1]{#1}
\csname url@samestyle\endcsname
\providecommand{\newblock}{\relax}
\providecommand{\bibinfo}[2]{#2}
\providecommand{\BIBentrySTDinterwordspacing}{\spaceskip=0pt\relax}
\providecommand{\BIBentryALTinterwordstretchfactor}{4}
\providecommand{\BIBentryALTinterwordspacing}{\spaceskip=\fontdimen2\font plus
\BIBentryALTinterwordstretchfactor\fontdimen3\font minus
  \fontdimen4\font\relax}
\providecommand{\BIBforeignlanguage}[2]{{%
\expandafter\ifx\csname l@#1\endcsname\relax
\typeout{** WARNING: IEEEtran.bst: No hyphenation pattern has been}%
\typeout{** loaded for the language `#1'. Using the pattern for}%
\typeout{** the default language instead.}%
\else
\language=\csname l@#1\endcsname
\fi
#2}}
\providecommand{\BIBdecl}{\relax}
\BIBdecl

\bibitem{fragapane2020increasing}
G.~Fragapane, D.~Ivanov, M.~Peron, F.~Sgarbossa, and J.~O. Strandhagen,
  ``Increasing flexibility and productivity in industry 4.0 production networks
  with autonomous mobile robots and smart intralogistics,'' \emph{Annals of
  operations research}, pp. 1--19, 2020.

\bibitem{alatise2020review}
M.~B. Alatise and G.~P. Hancke, ``A review on challenges of autonomous mobile
  robot and sensor fusion methods,'' \emph{IEEE Access}, vol.~8, pp.
  39\,830--39\,846, 2020.

\bibitem{faust2018prm}
A.~Faust, K.~Oslund, O.~Ramirez, A.~Francis, L.~Tapia, M.~Fiser, and
  J.~Davidson, ``Prm-rl: Long-range robotic navigation tasks by combining
  reinforcement learning and sampling-based planning,'' in \emph{2018 IEEE
  International Conference on Robotics and Automation (ICRA)}.\hskip 1em plus
  0.5em minus 0.4em\relax IEEE, 2018, pp. 5113--5120.

\bibitem{chiang2019learning}
H.-T.~L. Chiang, A.~Faust, M.~Fiser, and A.~Francis, ``Learning navigation
  behaviors end-to-end with autorl,'' \emph{IEEE Robotics and Automation
  Letters}, vol.~4, no.~2, pp. 2007--2014, 2019.

\bibitem{mnih2015human}
V.~Mnih, K.~Kavukcuoglu, D.~Silver, A.~A. Rusu, J.~Veness, M.~G. Bellemare,
  A.~Graves, M.~Riedmiller, A.~K. Fidjeland, G.~Ostrovski \emph{et~al.},
  ``Human-level control through deep reinforcement learning,'' \emph{nature},
  vol. 518, no. 7540, pp. 529--533, 2015.

\bibitem{dugas2020navrep}
D.~Dugas, J.~Nieto, R.~Siegwart, and J.~J. Chung, ``Navrep: Unsupervised
  representations for reinforcement learning of robot navigation in dynamic
  human environments,'' \emph{arXiv preprint arXiv:2012.04406}, 2020.

\bibitem{chen2017decentralized}
Y.~F. Chen, M.~Liu, M.~Everett, and J.~P. How, ``Decentralized
  non-communicating multiagent collision avoidance with deep reinforcement
  learning,'' in \emph{2017 IEEE international conference on robotics and
  automation (ICRA)}.\hskip 1em plus 0.5em minus 0.4em\relax IEEE, 2017, pp.
  285--292.

\bibitem{chen2017socially}
Y.~F. Chen, M.~Everett, M.~Liu, and J.~P. How, ``Socially aware motion planning
  with deep reinforcement learning,'' in \emph{2017 IEEE/RSJ International
  Conference on Intelligent Robots and Systems (IROS)}.\hskip 1em plus 0.5em
  minus 0.4em\relax IEEE, 2017, pp. 1343--1350.

\bibitem{everett2018motion}
M.~Everett, Y.~F. Chen, and J.~P. How, ``Motion planning among dynamic,
  decision-making agents with deep reinforcement learning,'' in \emph{2018
  IEEE/RSJ International Conference on Intelligent Robots and Systems
  (IROS)}.\hskip 1em plus 0.5em minus 0.4em\relax IEEE, 2018, pp. 3052--3059.

\bibitem{hochreiter1997long}
S.~Hochreiter and J.~Schmidhuber, ``Long short-term memory,'' \emph{Neural
  computation}, vol.~9, no.~8, pp. 1735--1780, 1997.

\bibitem{chen2020learning}
C.~Chen, S.~Majumder, Z.~Al-Halah, R.~Gao, S.~K. Ramakrishnan, and K.~Grauman,
  ``Learning to set waypoints for audio-visual navigation,'' \emph{arXiv
  preprint arXiv:2008.09622}, vol.~1, no.~2, p.~6, 2020.

\bibitem{chen2020soundspaces}
C.~Chen, U.~Jain, C.~Schissler, S.~V.~A. Gari, Z.~Al-Halah, V.~K. Ithapu,
  P.~Robinson, and K.~Grauman, ``Soundspaces: Audio-visual navigation in 3d
  environments,'' in \emph{Proceedings of the European Conference on Computer
  Vision (ECCV)}.\hskip 1em plus 0.5em minus 0.4em\relax Springer, 2020.

\bibitem{bansal2020combining}
S.~Bansal, V.~Tolani, S.~Gupta, J.~Malik, and C.~Tomlin, ``Combining optimal
  control and learning for visual navigation in novel environments,'' in
  \emph{Conference on Robot Learning}.\hskip 1em plus 0.5em minus 0.4em\relax
  PMLR, 2020, pp. 420--429.

\bibitem{guldenringlearning}
R.~G{\"u}ldenring, M.~G{\"o}rner, N.~Hendrich, N.~J. Jacobsen, and J.~Zhang,
  ``Learning local planners for human-aware navigation in indoor
  environments.''

\bibitem{regier2020deep}
P.~Regier, L.~Gesing, and M.~Bennewitz, ``Deep reinforcement learning for
  navigation in cluttered environments,'' 2020.

\bibitem{kinoastar}
B.~Zhou, F.~Gao, L.~Wang, C.~Liu, and S.~Shen, ``Robust and efficient quadrotor
  trajectory generation for fast autonomous flight,'' \emph{IEEE Robotics and
  Automation Letters}, vol.~4, no.~4, pp. 3529--3536, 2019.

\bibitem{deboor}
K.~Qin, ``General matrix representations for b-splines,'' \emph{The Visual
  Computer}, vol.~16, no. 3-4, pp. 177--186, 2000.

\bibitem{egoplanner}
X.~Zhou, Z.~Wang, H.~Ye, C.~Xu, and F.~Gao, ``Ego-planner: An esdf-free
  gradient-based local planner for quadrotors,'' \emph{IEEE Robotics and
  Automation Letters}, 2020.

\bibitem{rosmann2015timed}
C.~R{\"o}smann, F.~Hoffmann, and T.~Bertram, ``Timed-elastic-bands for
  time-optimal point-to-point nonlinear model predictive control,'' in
  \emph{2015 european control conference (ECC)}.\hskip 1em plus 0.5em minus
  0.4em\relax IEEE, 2015, pp. 3352--3357.

\bibitem{rosmann2019time}
C.~R{\"o}smann, ``Time-optimal nonlinear model predictive control,'' Ph.D.
  dissertation, Dissertation, Technische Universit{\"a}t Dortmund, 2019.

\bibitem{kastner2020deep}
L.~K{\"a}stner, C.~Marx, and J.~Lambrecht, ``Deep-reinforcement-learning-based
  semantic navigation of mobile robots in dynamic environments,'' in \emph{2020
  IEEE 16th International Conference on Automation Science and Engineering
  (CASE)}.\hskip 1em plus 0.5em minus 0.4em\relax IEEE, 2020, pp. 1110--1115.

\end{thebibliography}

\end{document}